\documentclass[a4paper, 10pt, conference]{ieeeconf}      
\usepackage{FG2021}
\usepackage{hyperref}
\usepackage{graphicx}
\usepackage{subcaption}
\usepackage{xcolor}
\newcommand\crule[3][black]{\textcolor{#1}{\rule{#2}{#3}}}
\FGfinalcopy 

\IEEEoverridecommandlockouts                              
\def\FGPaperID{****} 

\title{\LARGE \bf
Does Melania Trump have a body double from the perspective of automatic face recognition?
}

\author{\parbox{16cm}{\centering
    {\large Khawla Mallat$^1$, Fabiola Becerra-Riera$^2$, Annette Morales-Gonz\'alez$^2$, Heydi M\'endez-V\'azquez$^2$ and Jean-Luc Dugelay$^1$}\\
    {\normalsize
    $^1$ Digital Security Department, EURECOM Campus Sophia Tech, 450 route des Chappes F-06410 Biot Sophia Antipolis, France. Email: \{mallat, dugelay\}@eurecom.fr\\
    $^2$Advanced Technologies Application Center (CENATAV) 7A \#21406 Siboney, Playa, P.C.12200, Havana, Cuba. Email: \{fbecerra,amorales,hmendez\}@cenatav.co.cu}}
}

\begin{document}

\ifFGfinal
\thispagestyle{empty}
\pagestyle{empty}
\else
\author{Anonymous FG2021 submission\\ Paper ID \FGPaperID \\}
\pagestyle{plain}
\fi
\maketitle

\begin{abstract}

In this paper, we explore whether automatic face recognition can help in verifying widespread misinformation on social media, particularly conspiracy theories that are based on the existence of body doubles. The conspiracy theory addressed in this paper is the case of the Melania Trump body double. We employed four different state-of-the-art descriptors for face recognition to verify the integrity of the claim of the studied conspiracy theory. In addition, we assessed the impact of different image quality metrics on the variation of face recognition results. Two sets of image quality metrics were considered: acquisition-related metrics and subject-related metrics.  

\end{abstract}

\section{Introduction}

This is not Melania! In October 2017, the hypothesis of the existence of a body double of Melania TRUMP appeared for the first time on social networks. Then regularly, what is now referred to as \emph{`The Melania Trump replacement conspiracy theory'}\footnote{ \href{https://en.wikipedia.org/wiki/Melania\_Trump\_replacement\_conspiracy\_theory}{https://en.wikipedia.org/wiki/Melania\_Trump\_replacement\_conspiracy\_theory}} by mass media, reappears in the front of the stage, at least once a year, with a peak during the last presidential election campaign.
Face is used by all of us as the main human trait to recognize each other in daily life. This is why, even if some other criteria have been used to demonstrate that Melania is not Melania, like her attitude and behavior or her body height, most arguments advanced by people are related to her facial appearance. 

Algorithms based on artificial intelligence, image/video processing and computer vision are designed and used to fight against fake news that include doctored images~\cite{mahfoudi2019defacto} and/or AI-based generated videos (i.e. deepfakes)~\cite{dolhansky2020deepfake}. The issue addressed in this article is different in the sense that the open question is not about the integrity of the picture but about the physical presence of the real Melania or of a body double in the scene recorded by the camera. So Biometrics (and more particularly Automatic Face Recognition), more than image forensic tools, is the appropriate technology to address the problem.

 There are two possibilities in case of the non-existence of a body double of Melania: either people are complotists in the sense that they are aware that Melania has no double body but claim the contrary or, people are of good faith and consider by visual inspection that there are two different people on photographs: Melania and a body double based on some possible visual differences among faces. Visual inspection by human is subjective and today less accurate than a digital and objective inspection by machines for many tasks related to face analysis~\cite{PHILLIPS201474}. This is why automatic face recognition may significantly help in the process for forming a robust opinion concerning this supposed conspiracy theory.

In this work, we use automatic face recognition to conduct an analysis of a group of images of Melania Trump, in order to try to determine if she has a body double or not.
\section{Automatic face recognition}

In this section, we present primarily the image samples of Melania Trump on which our study has been based. Then, we introduce the state-of-the-art descriptors of face recognition employed in this study followed by the experimental protocol and results that will help answering the question \emph{`Does Melania Trump has a body double?'}.

\subsection{Face samples} \label{sec13}

\begin{figure}[ht]
\begin{center}
\includegraphics[width=\linewidth]{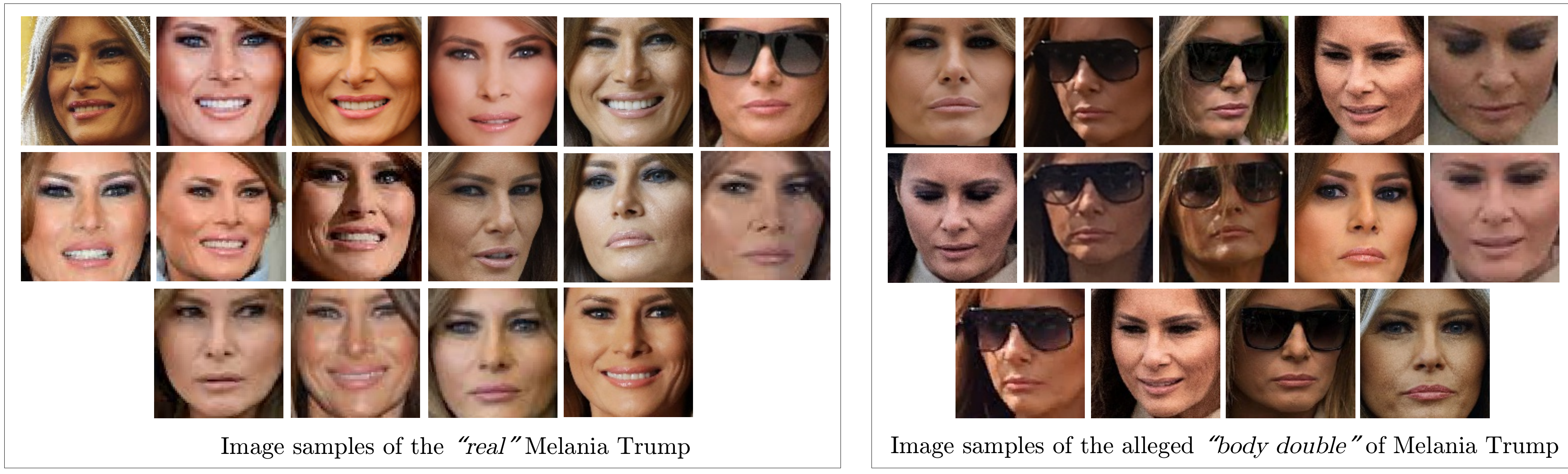}
\end{center}
   \caption{Preprocessed face images of the \emph{`real'} Melania and the alleged \emph{`body double'}. }
\label{fig:melania}
\vspace{-0.5cm}
\end{figure}

The presented study was driven by the interest of W9, a French TV channel, in the conspiracy theory case related to Melania Trump~\footnote{\href{https://www.youtube.com/watch?v=9HJKKYeJdTk&t=1239s}{https://www.youtube.com/watch?v=9HJKKYeJdTk\&t=1239s}}. The TV channel provided us with a set of images of Melania Trump collected from social media, mainly Twitter, collected by journalists. These images were labeled by the social media users as the \emph{`real'} Melania Trump or as the \emph{`body double'}. Faces were detected using RetinaFace~\cite{deng2020retinaface} and only faces of Melania Trump were selected and then aligned. Figure~\ref{fig:melania} shows the preprocessed face images. At first glance, one could see that there is an important discrepancy between the two sets of images. First, the set of the alleged \emph{`body double'} contains 6 out of 14 images in which Melania is wearing large dark glasses which covers up the ocular area, while it is the case for only 1 out of 16 images for the \emph{`real'} Melania. In addition, one could see that the set of images of the \emph{`real'} Melania contains images where Melania is captured with a smiling face, open eyes and proper illumination conditions, while the images contained in the set of the alleged \emph{`body double'} of Melania seem to be taken from video recordings, where Melania appears with blinking eyes, head down and low illumination conditions.

\subsection{Automatic face recognition models}
In this study, we employ four state-of-the-art face recognition descriptors that achieved more than 99.3\% of accuracy when evaluated on LFW dataset. For each face descriptor, feature embeddings are extracted from Melania's face images and compared against each other using cosine similarity.

\textbf{LightCNN~\cite{wu2018light}} is an implementation of CNN for face recognition designed to have fewer trainable parameters than a vanilla CNN and to handle noisy labels. This network introduces a new concept of max-out activation in each convolutional layer for feature filter selection. We use the pretrained network with 29-layers to obtain embeddings of 256-dimension from face images. 

\textbf{MobileSqueezeNet~\cite{mobilesqueezenet}} is a lightweight CNN released with Intel's OpenVino framework and model zoo, for face recognition in re-identification scenarios. It is based on MobileNet V2~\cite{howard2017mobilenets} backbone, which consists of 3x3 inverted residual blocks with squeeze-excitation attention modules. After the backbone, the network applies global depthwise pooling and then uses 1x1 convolution to create the final 256-dimensional embedding vector.

\textbf{SphereFace~\cite{liu2017sphereface}} refers to angularly distributed feature embeddings designed for face recognition. In contrast to employing Euclidean margin to the CNN’s learned features, SphereFace proposes to use an angular softmax (named A-Softmax) to learn discriminative face features with a novel geometric interpretation. A-Softmax loss can be interpreted as constraining learned features to be discriminative on a hypersphere manifold, which intrinsically matches the prior that face images lie on a manifold. We use the 512-dimensional SphereFace embeddings.

\textbf{ArcFace~\cite{deng2019arcface}} is Additive Angular Margin loss function, a new loss function for face recognition designed to improve the discriminative power of the learnt feature embeddings. The proposed loss function aims to optimise the geodesic distance margin by considering the correspondence between the angle and arc in the normalised hypersphere. We used a model pretrained on LFW dataset to extract embeddings of 512-dimension from face images.

\subsection{Experimental protocol and results}\label{sec3.3}

\begin{figure*}[ht]
\centering
\begin{subfigure}{0.225\linewidth}
  \centering
  \includegraphics[width=\linewidth, trim={2cm 0.2cm 1.5cm 1cm}, clip ]{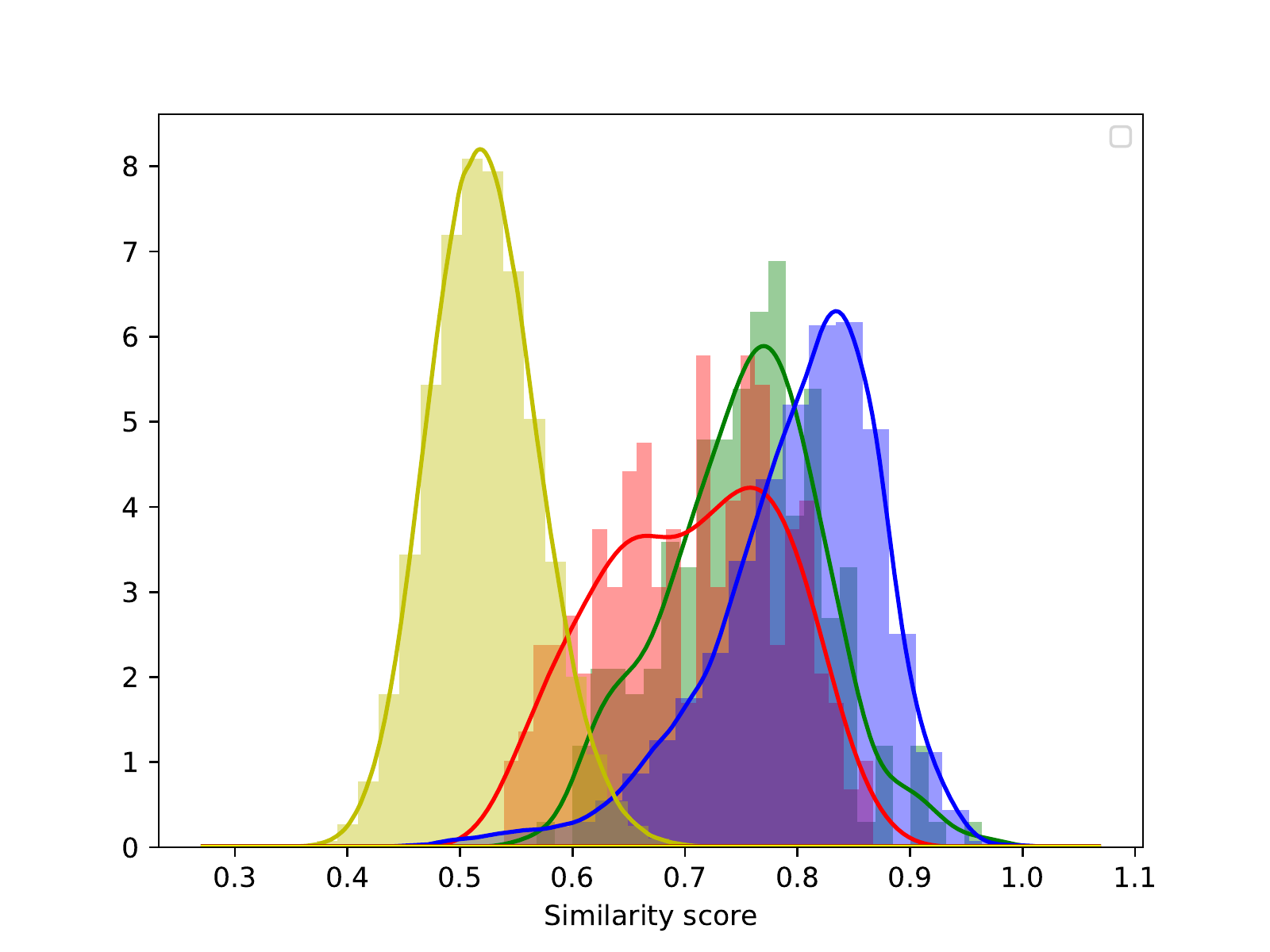}
  \caption{}
  \label{fig:ald}
\end{subfigure}%
\begin{subfigure}{0.225\linewidth}
  \centering
  \includegraphics[width=\linewidth, trim={2cm 0.2cm 1.5cm 1cm}, clip ]{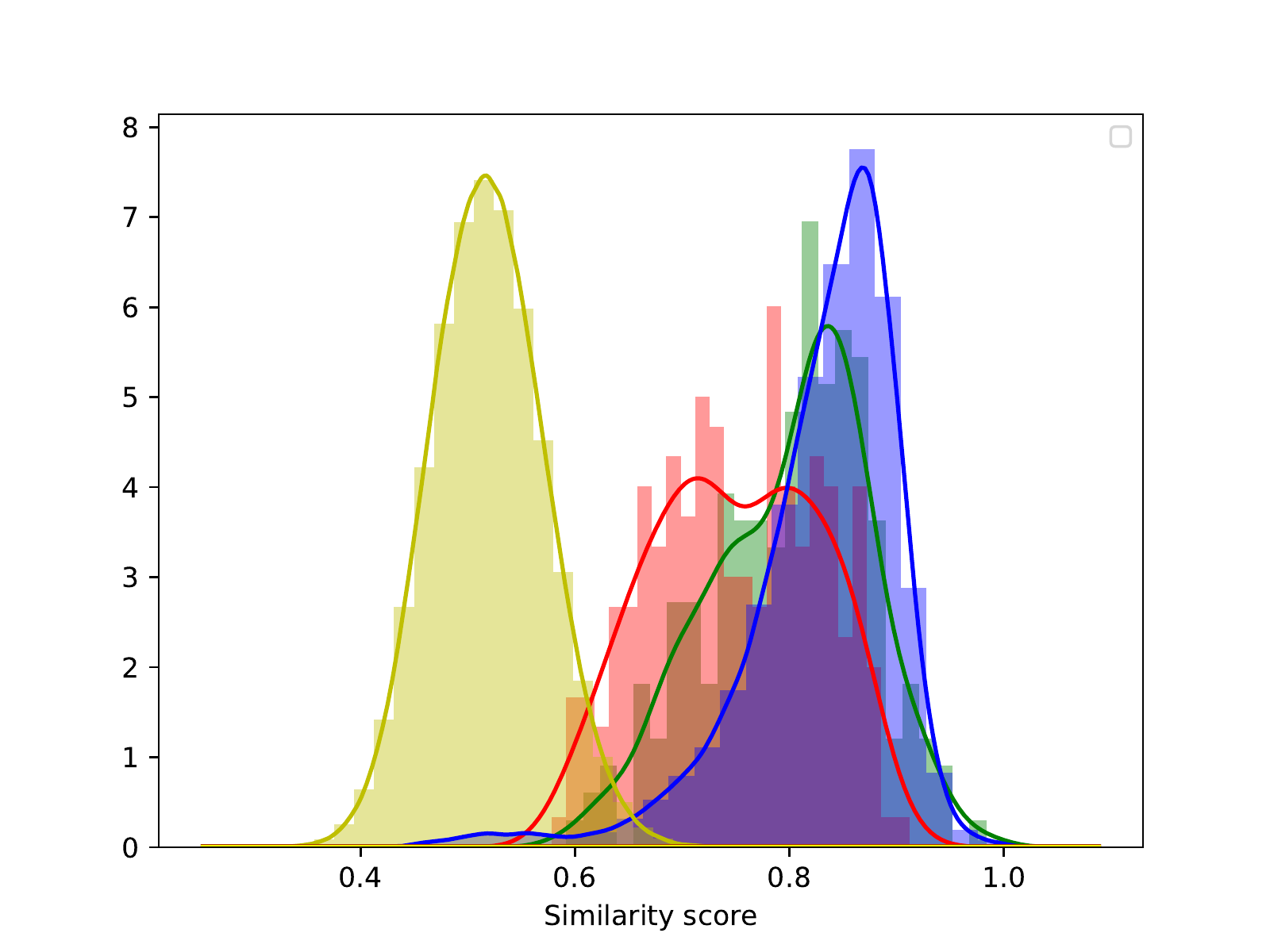}
  \caption{}
  \label{fig:bld}
\end{subfigure}%
\begin{subfigure}{0.225\linewidth}
  \centering
  \includegraphics[width=\linewidth, trim={2cm 0.2cm 1.5cm 1cm}, clip ]{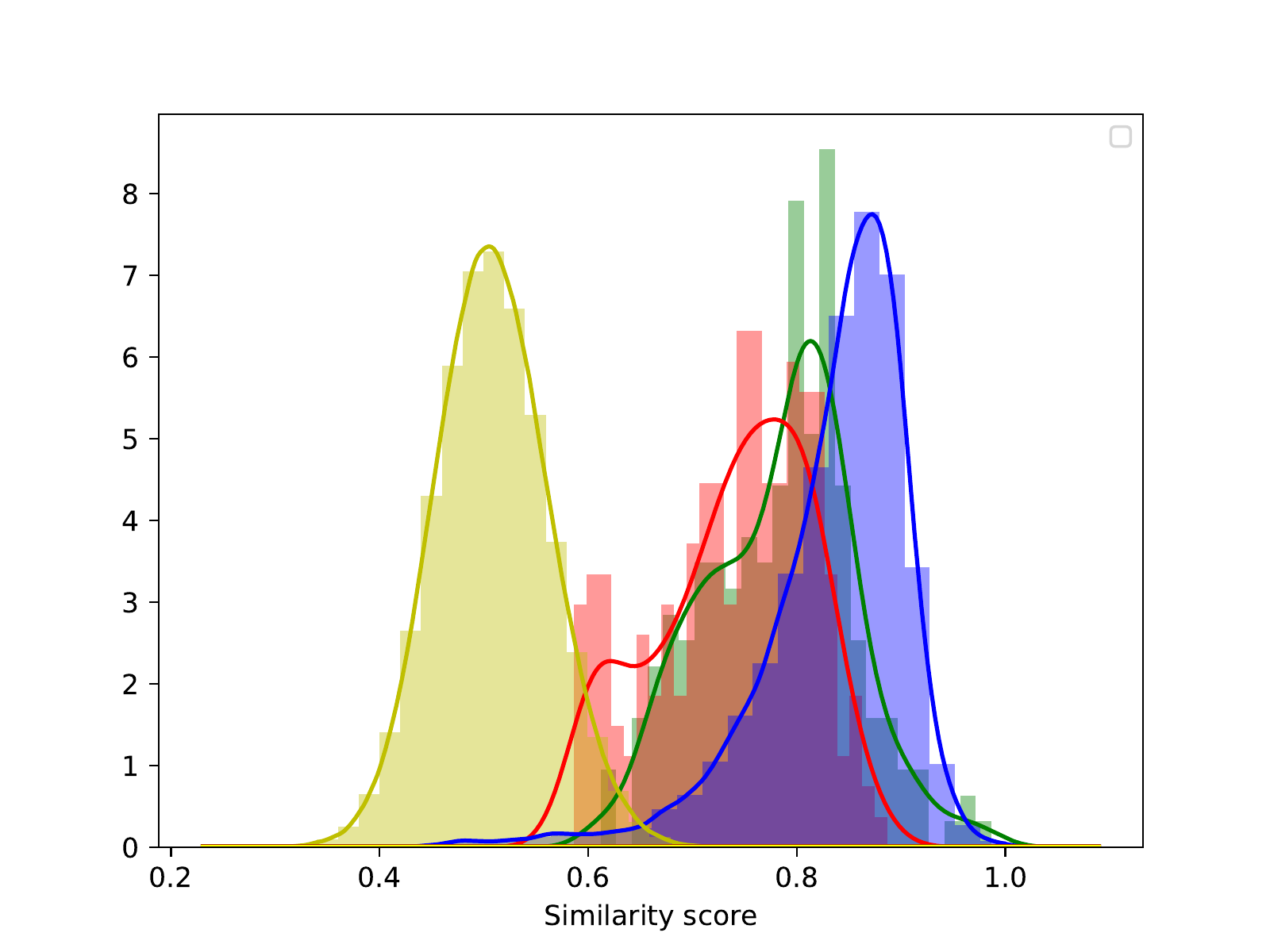}
  \caption{}
  \label{fig:cld}
\end{subfigure}%
\begin{subfigure}{0.225\linewidth}
  \centering
  \includegraphics[width=\linewidth, trim={2cm 0.2cm 1.5cm 1cm}, clip ]{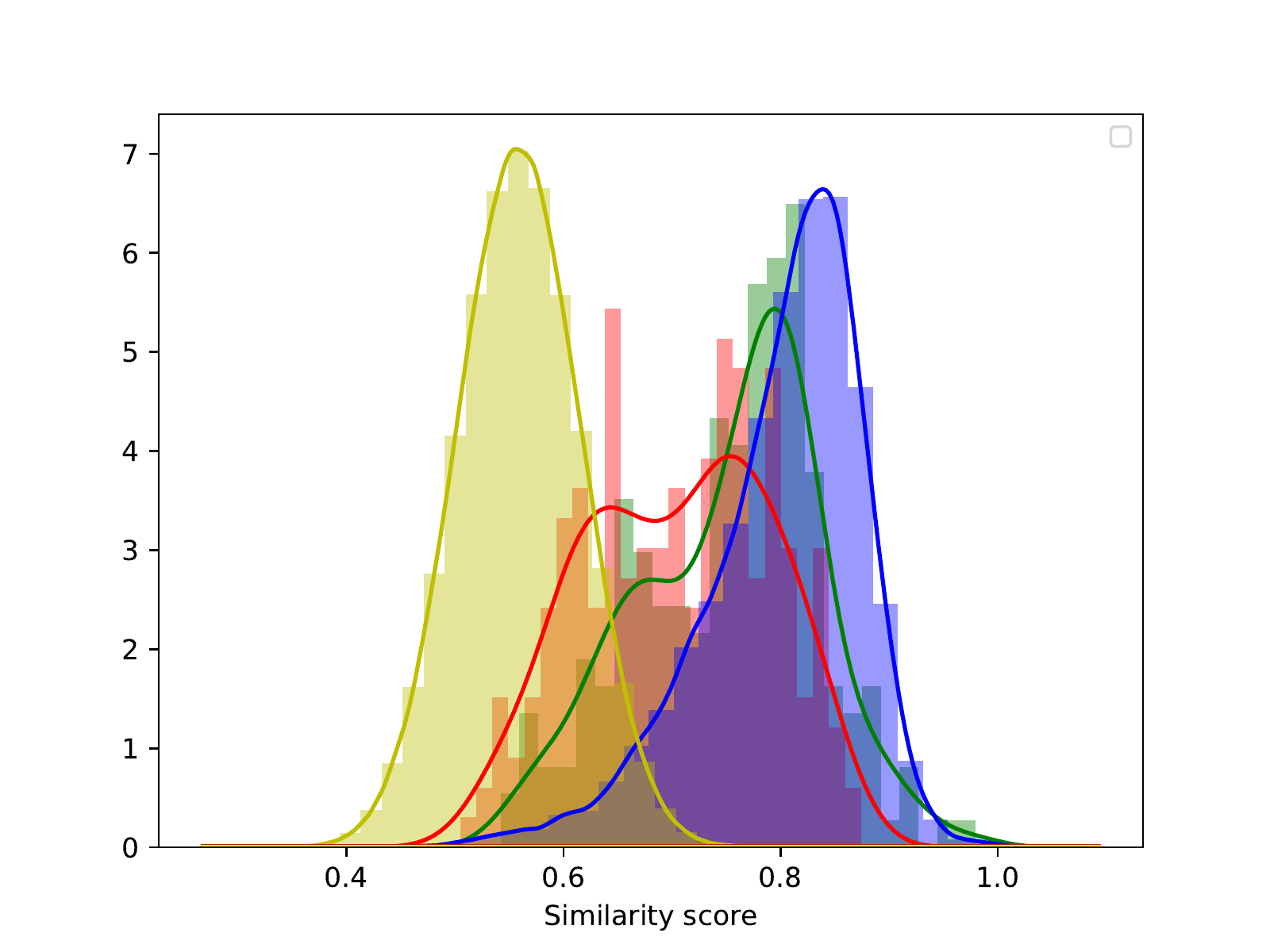}
  \caption{}
  \label{fig:dld}
\end{subfigure}%
\newline
\crule[green!50!black!50]{0.25cm}{0.25cm} Genuines (Melania) \crule[red!100!black!50]{0.25cm}{0.25cm} Imposters (Melania) \crule[blue!100!black!50]{0.25cm}{0.25cm} Genuines (Celebrities) \crule[yellow!100!black!50]{0.25cm}{0.25cm} Imposters( Celebrities)
\caption{Similarity score distribution of \emph{`real'} against \emph{`body double'} Melania Trump (a) Arcface (b) LightCNN (c) MobileSqueezeNet (d) SphereFace.}
\label{fig:allVSall}
\vspace{-0.5cm}
\end{figure*}

Assuming we have 2 identities for Melania Trump, the first consists of the \emph{`real'} Melania and the second consists of the alleged \emph{`body double'}, we performed a face verification experiment using an All-vs-All comparison scheme. \emph{Genuine} pairs belong to the same identity, while \emph{Impostors} pairs correspond to the comparison of one identity vs. the other, meaning the \emph{`real'} Melania against the alleged \emph{`body double'}. We illustrate in Figure~\ref{fig:allVSall} the score distributions of Genuines in green and of Impostors in red using the four selected face descriptors. We also introduced the score distributions of imposters and genuines deduced from a dataset, collected from the web by the authors, of female celebrities in their 50s similar in profile to Melania Trump. This dataset contains 50 identities, with 30 images per identity.

It can be seen that the genuine (in green) and the imposter (in red) score distributions overlap substantially, although, there is a slight difference between the two score distributions. This difference might be due to the discrepancy in image samples between the set of \emph{`real'} Melania and the alleged \emph{`body double'} as depicted in section~\ref{sec13}. In section~\ref{quality}, we evaluate the impact of different factors that might affect the automatic face recognition score as an attempt to justify the slight difference between the green and the red score distribution.
In addition, one could see that the distributions of genuine and imposter scores deduced from the Melania data generally fall into the genuine score distribution deduced from the celebrities dataset, with a minor overlap with the imposters score distribution, meaning that the images of the real Melania and the images of the alleged body double belong to the same person.

\section{Factors affecting automatic face recognition} \label{quality}
In this section, we suppose that Melania Trump does not have a body double according to the significant overlap between genuine and imposter score distributions deduced from Melania image set, illustrated in section~\ref{sec3.3}. However, in this section, we attempt to justify the difference between these two distributions highlighted hereabove.
There are several factors that could affect the utility of a face image for a specific task. In this section, since we are particularly interested in face recognition, we explored the impact of key quality metrics~\cite{mendez2012evaluacion, becerra2021attribute, fusionMallat} which were considered according to the particular conditions of our case of study and that are subject and environmental/sensor-related.

To highlight the impact of a given quality metric on the face recognition performance, we sorted all the 30 image samples of Melania Trump according to their corresponding quality measure in an ascending order. Feature embeddings were extracted from all the images. Then, the affinity matrix of the cosine similarity scores is computed between feature embeddings of Melania Trump's images in an All-vs-All comparison scheme. Hereafter, only results obtained using ArcFace feature embeddings will be reported as all the four face descriptors depict the same behaviour.

\subsection{Acquisition-related image quality metrics}
From simply looking at the image samples of Melania Trump, one could see a prominent difference between the \emph{`real'} and the alleged \emph{`body double'} sets mainly in the image acquisition quality. 

\textbf{Brightness}~\cite{fusionMallat}
Depending on the acquisition conditions and the camera devices, images could often be affected by variations in terms of illumination and color leveling (the degree of lightness or darkness of a color) which result in brightness variations. Brightness is the perception caused by the luminance of an object, a face in our case.
Major fluctuations of these measures could drastically change the appearance of subjects causing different skin color and shaded or extremely brighten zones that could even partially occlude discriminative face parts. Since brightness does not visually appear to be homogeneous across the Melania Trump image set, we decided to analyse how much it really affects the recognition.

\textbf{Face luminance}~\cite{mendez2012evaluacion}
Face luminance deals with the distribution of light across the face and it is measured in specific face regions that are prominent for the face recognition process.
The face image quality based on luminance, was measured by averaging the normalized mean luminance values for a set of triangles, representing key face regions, that are more likely to be affected by changes in illumination. Since illumination does not appear to be homogeneously distributed across the face, an assessment of the variation of face luminance on the face recognition score is conducted.

\textbf{Exposure}~\cite{fusionMallat} is the amount of light that reaches the acquisition sensor and it can be controlled using aperture and shutter settings. When the face image is correctly exposed, the image provides rich details in both dark shadow regions as well as bright regions.
The Melania Trump set of images contain mainly underexposed images in which the dark shadow regions lose fine details and appear as pure block regions because of receiving too little light.

\textbf{Contrast}~\cite{fusionMallat}
Low contrast images are commonly due to poor illumination conditions during the image acquisition, wrong setting of the aperture and shutter speed. Low contrast face images contain underexposed and overexposed regions that result in loss of geometrical and color information of facial attributes. Image contrast is defined as the separation between the darkest and brightest areas of the image, and it can be computed as the difference in color intensities.

\textbf{Sharpness}~\cite{mendez2012evaluacion}
In blurred 
images, the face regions (eyes, nose, mouth, etc.) are usually not sufficiently clear or well delimited due to the loss of visual information by confusing and mixing shapes, textures and colors. To decide to which extent more or less defined/sharp edges are influencing the face recognition process, we applied an unsharp masking process to sharpen the original face image and average the resulting pixel intensities to obtain a sharpness quality score.

\subsection{Subject-related image quality metrics}
The second main difference between the set of the \emph{`real'} and the alleged \emph{`body double'} is related to the physical appearance of Melania herself. One could see that the set of the \emph{`real'} Melania shows images of her with the head straight and a glowy face, whilst for the set of the alleged \emph{`body double'}, Melania appears either wearing sunglasses or with the eyes closed with a varying head pose. Accordingly, 3 quality metrics that are subject-related were selected to evaluate their impact on face recognition similarity score.

\textbf{Occlusion by sunglasses}~\cite{mendez2012evaluacion}
Since several samples of the alleged \emph{`body double'} set show Melania wearing sunglasses, our focus was drawn on the occlusion of the periocular region, which has been shown to be a very discriminative region in face recognition, and its impact on face recognition performance. Therefore, to measure the quality of a face image based on the use of sunglasses, we first used a CNN classifier as sunglasses detector, and then we employed the output probability of the class representing “non sunglasses” as final quality score: face images with higher `non sunglasses' probability are more likely to be good quality images and vice versa.

\textbf{Femininity}~\cite{femininity}
A main factor that might have affected the face recognition performance can clearly observed in the Melania Trump images is her physical appearance. In the set of the \emph{`real'} Melania, she appears with radiant face and make up on, whilst in the alleged \emph{`body double'} set, Melania seems to be wearing too little make up or no make up at all, with a wearied face. Consequently, we propose to quantify the femininity appearance of Melania Trump on all the image samples and evaluate their impact on face recognition performance. For this purpose, a gender estimation model~\cite{antipov} was employed to deliver a score representing the probability of being a male or a female. We consider the femininity score as the probability of being a female. Mallat et al.~\cite{femininity} proved that the more makeup is applied (up to a certain level), the higher the femininity score is.

\textbf{Head pose}~\cite{mendez2012evaluacion}
 The complexity of dealing with multiple head rotations under uncontrolled conditions has long been a bottleneck in face recognition. Different head poses usually imply the occlusion or deformation of some face regions, which can lead to the mismatch of two samples of the same person just because they were captured from different angles. It is worth noting that general Melania postures are basically frontal and just a few samples present some yaw or pitch variation, in particular for the alleged \emph{“body double"} set. 
 
 \subsection{Discussion}
\begin{figure}[ht]
\centering
\begin{subfigure}{0.2\linewidth}
  \centering
  \includegraphics[width= 0.95\linewidth]{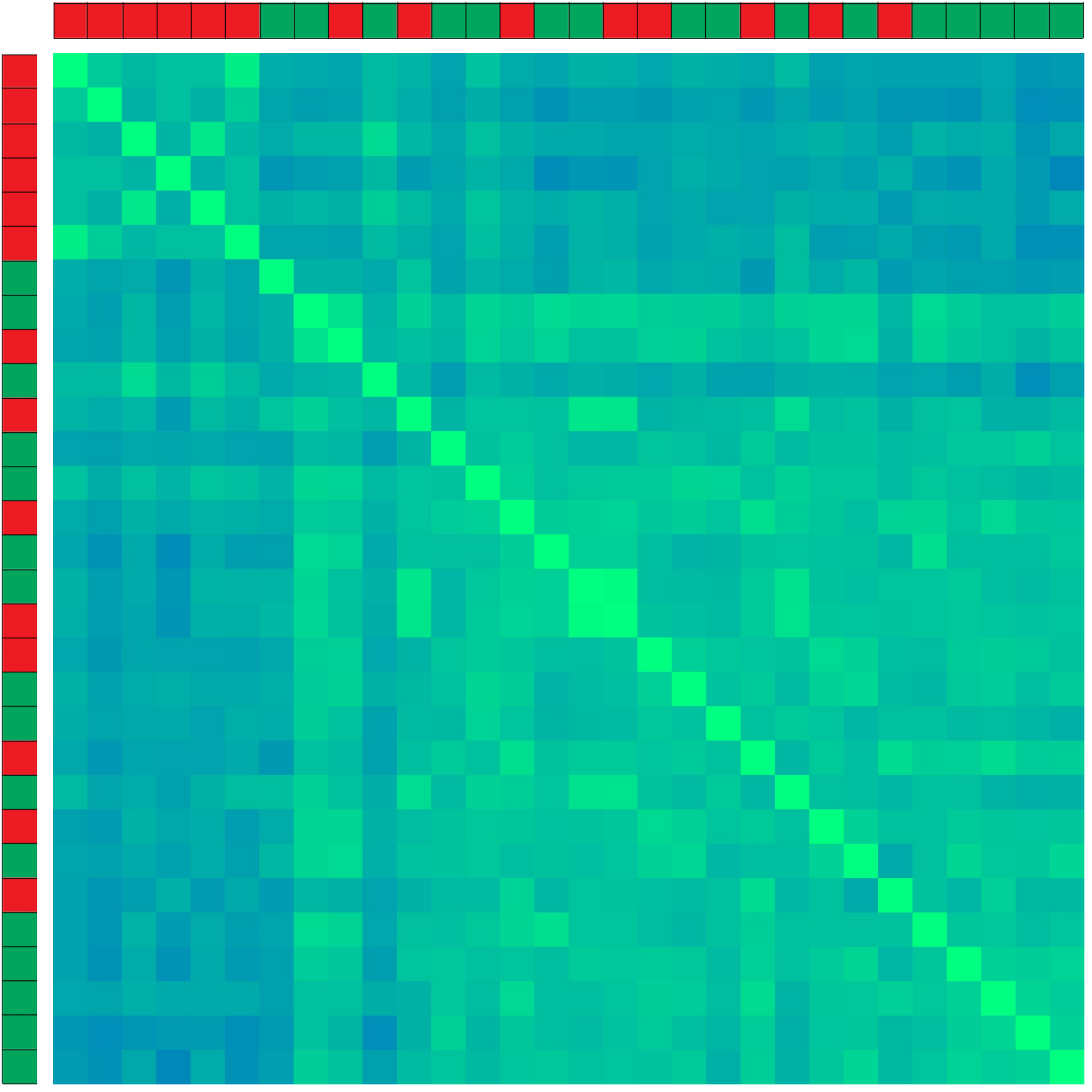}
  \caption{}
  \label{fig:brightness}
\end{subfigure}%
\begin{subfigure}{0.2\linewidth}
  \centering
  \includegraphics[width=0.95\linewidth]{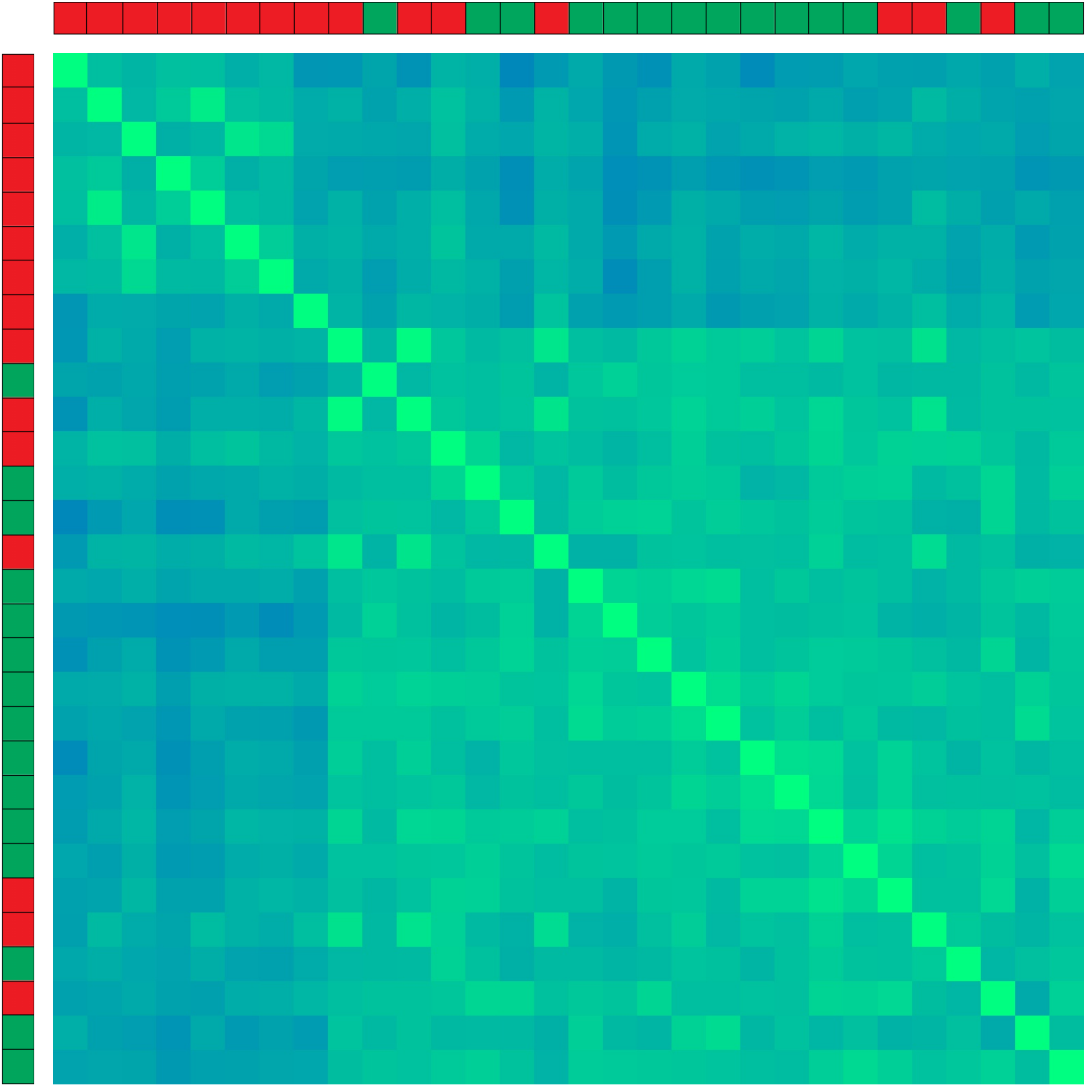}
  \caption{}
  \label{fig:luminance}
\end{subfigure}%
\begin{subfigure}{0.2\linewidth}
  \centering
  \includegraphics[width=0.95\linewidth ]{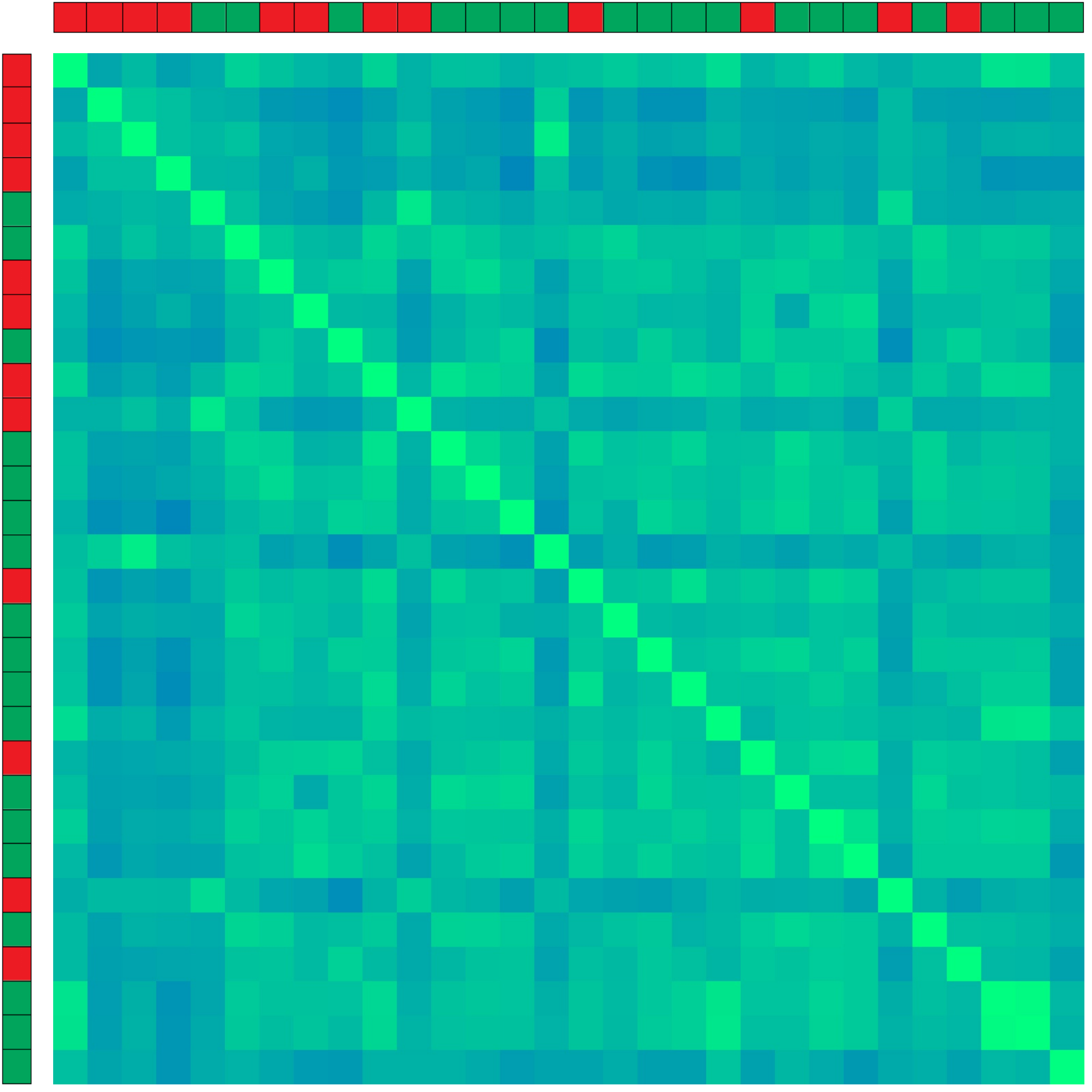}
  \caption{}
    \label{fig:exposure}
\end{subfigure}%
\begin{subfigure}{0.2\linewidth}
  \centering
  \includegraphics[width= 0.95\linewidth ]{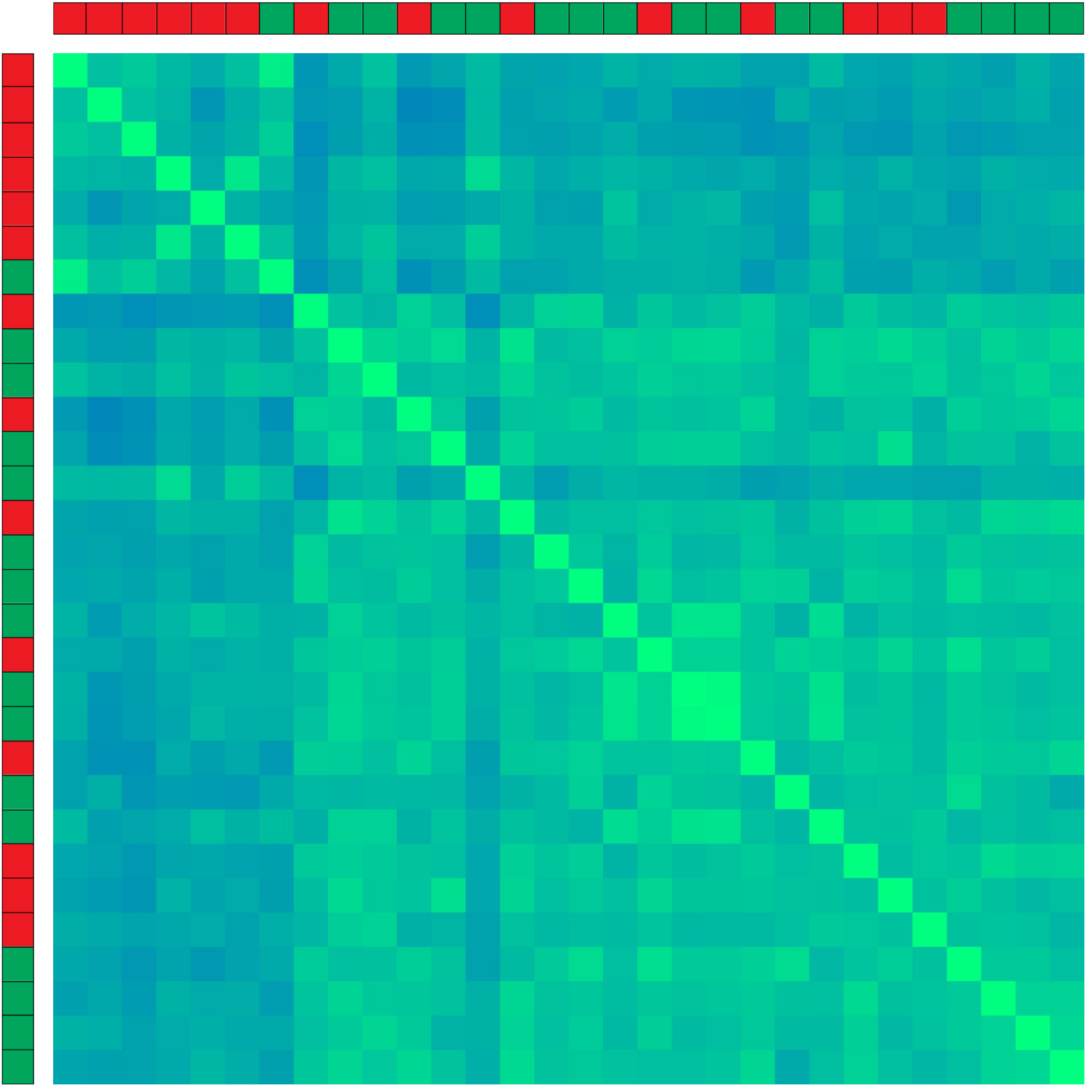}
  \caption{}
    \label{fig:contrast}
\end{subfigure}%
\begin{subfigure}{0.2\linewidth}
  \centering
  \includegraphics[width= 0.95\linewidth]{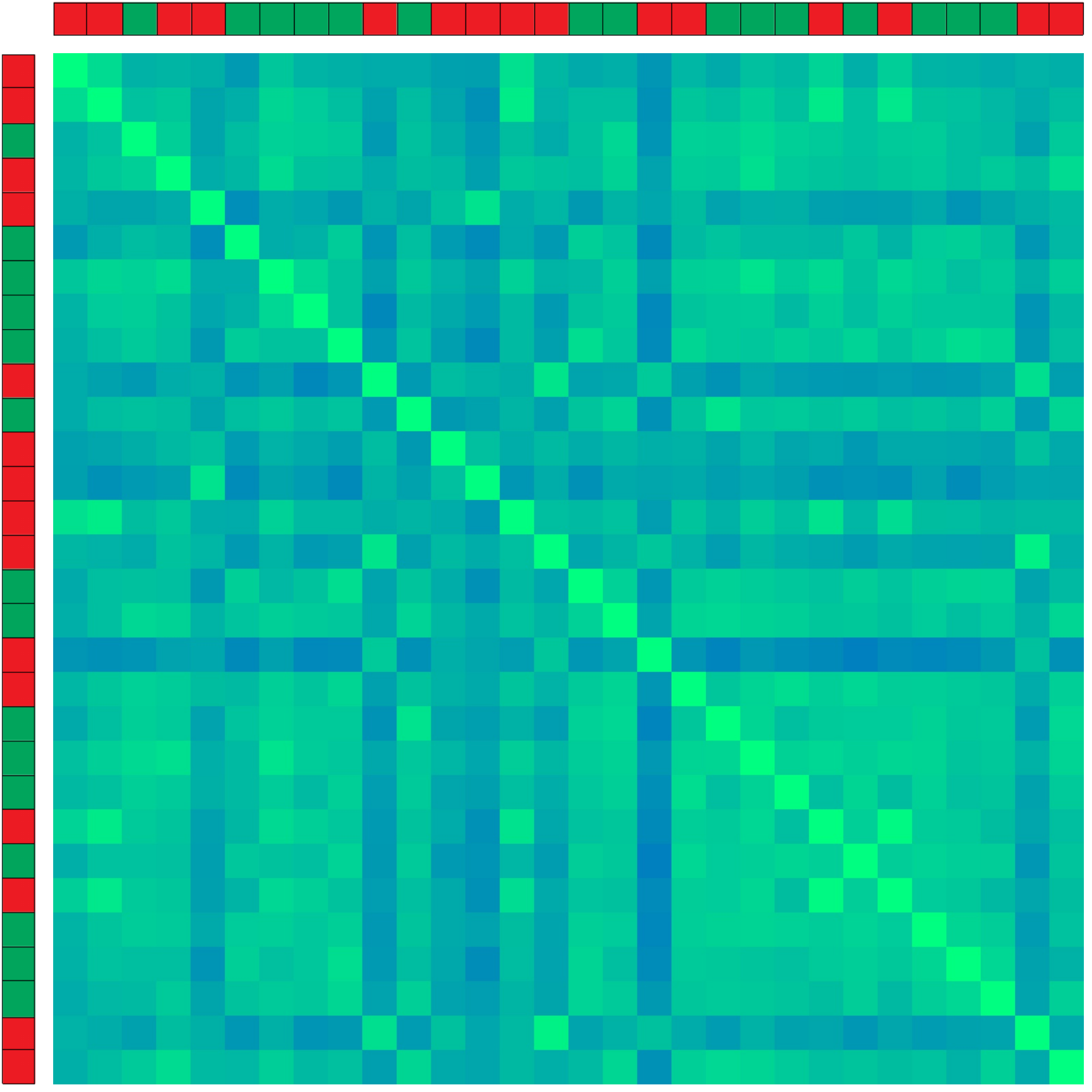}
  \caption{}
  \label{fig:sharpness}
\end{subfigure}%
\newline
 \vspace{0.1cm}
 \crule[green!50!black!100]{0.2cm}{0.2cm} \footnotesize	{image labeled as \emph{`real'} Melania} \hspace{0cm} \crule[red]{0.2cm}{0.2cm} \footnotesize	{image labeled as \emph{`body double'}}
  \vspace{0.1cm}
\caption{Affinity matrix of cosine similarity between ArcFace embeddings of Melania Trump's images, sorted according to: (a) Brightness (b) Face luminance (c) Contrast (d) Exposure (e) Sharpness. }
\label{fig:conventional_quality}
\vspace{-0.5cm}
\end{figure}

Figure~\ref{fig:conventional_quality} reports the affinity matrices of the cosine similarity scores that highlight the impact of image quality metrics that are related to the image acquisition: brightness, face luminance, sharpness, exposure and contrast. Blue cells reflect a low similarity score (closer to 0) whereas green cells correspond to high similarity score (close to 1). 

In Figures~\ref{fig:brightness},~\ref{fig:luminance},~\ref{fig:exposure} and ~\ref{fig:contrast}, one could see that images with low quality yield low similarity scores and they correspond to image samples from the alleged \emph{`body double'}, whilst high quality images deliver high similarity score and they are mostly associated to image samples of \emph{`real'} Melania. For the exposure quality metric, the lines of the affinity matrix, in Figure~\ref{fig:exposure}, that depict low similarity scores for images with higher exposure correspond to specific image samples in which Melania Trump appears either with sunglasses or with the head down. Figure~\ref{fig:sharpness} shows a small correlation between sharper images and high similarity scores. However, we can deduce from the matrix that there are different and more relevant factors that are influencing the face recognition performance. The cross-shaped blue area that depict low similarity scores from Figure~\ref{fig:sharpness} corresponds to the comparison of four images of average sharpness but considerably affected by illumination and using sunglasses; this is also the case of the two isolated sharp images producing the blue lines that converge in the bottom right corner of the matrix. 

\begin{figure}[ht]
\centering
\begin{subfigure}{0.3\linewidth}
  \centering
  \includegraphics[width=0.65\linewidth, ]{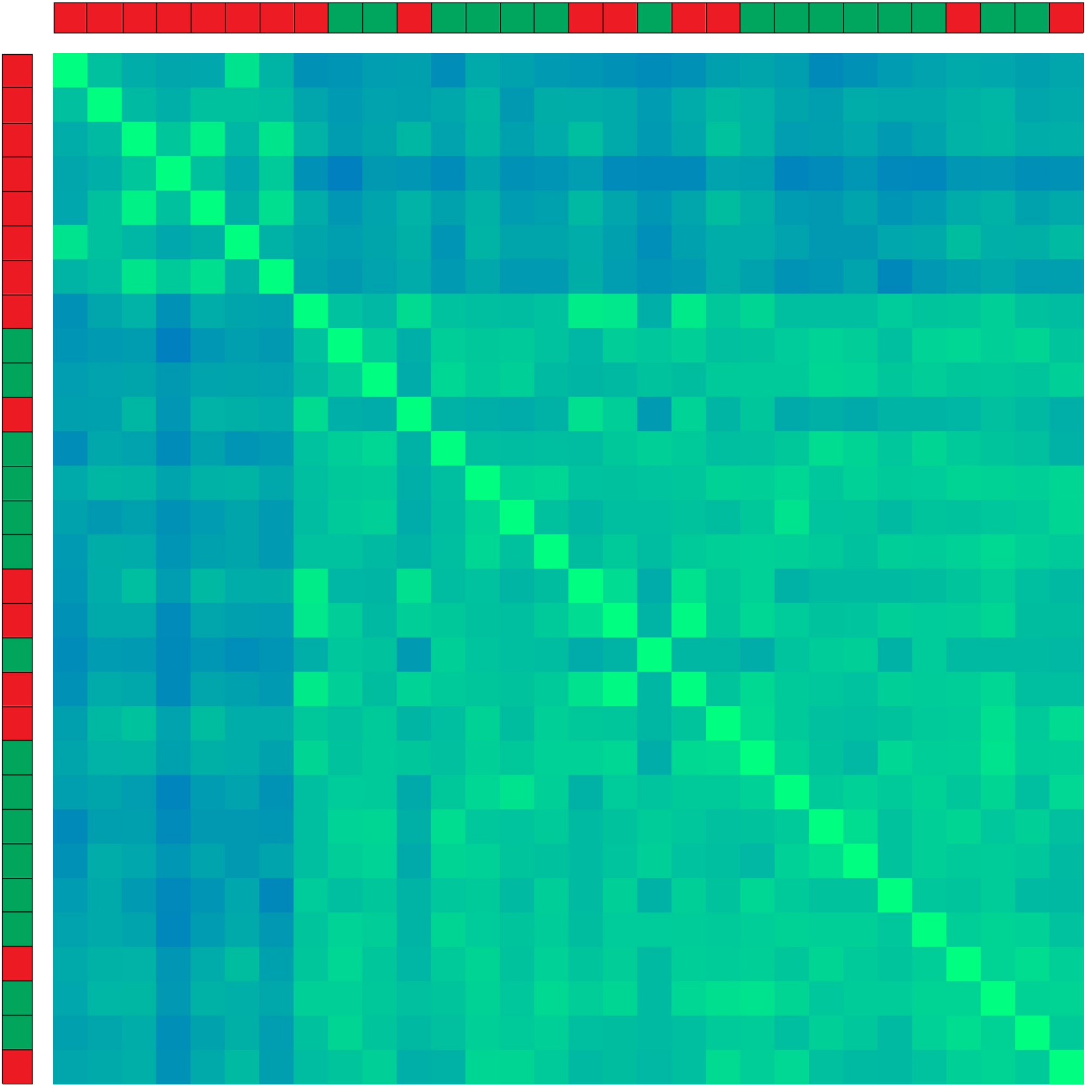}
  \caption{}
  \label{fig:sunglasses}
\end{subfigure}%
\begin{subfigure}{0.3\linewidth}
  \centering
  \includegraphics[width= 0.65\linewidth ]{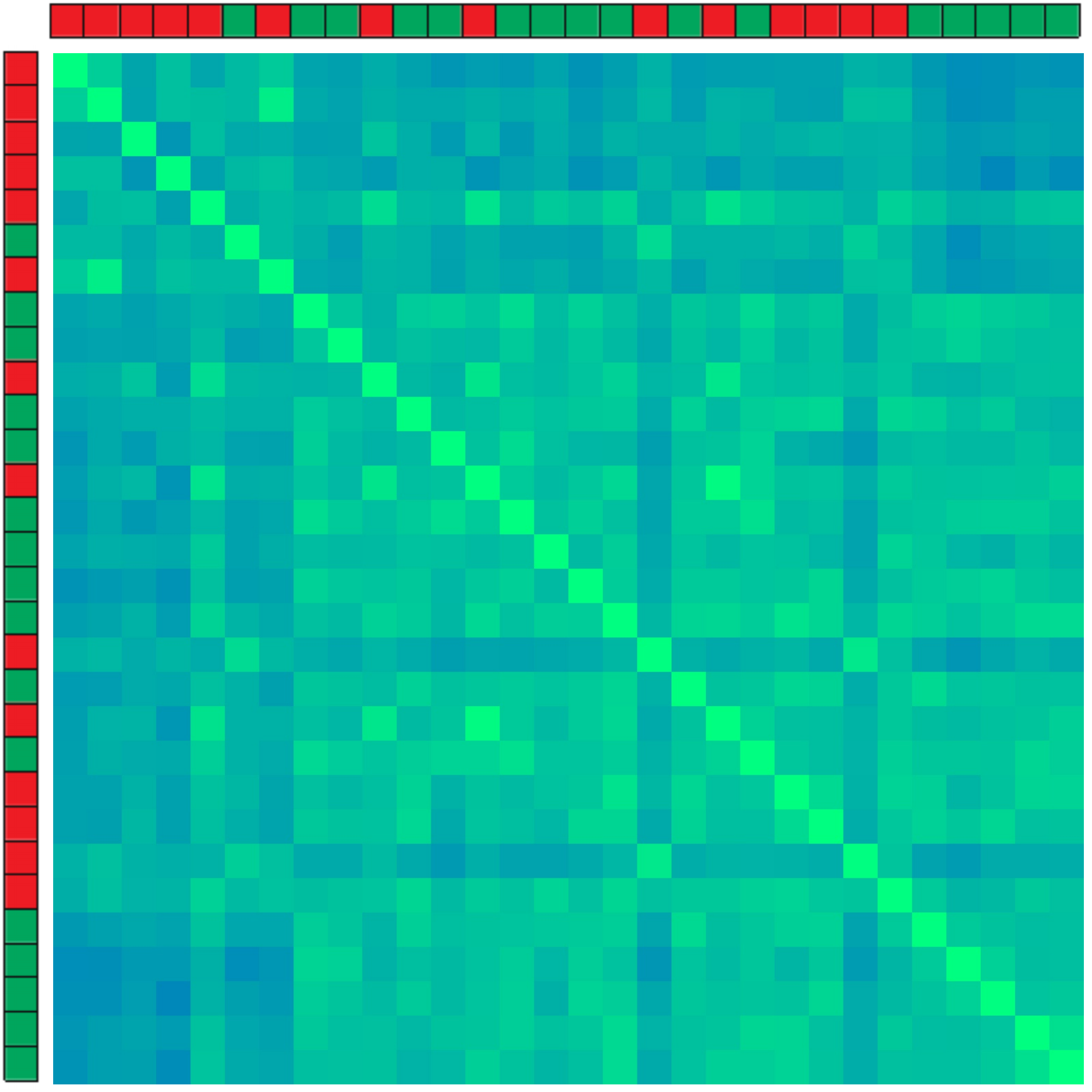}
  \caption{}
  \label{fig:beauty}
\end{subfigure}%
\begin{subfigure}{0.3\linewidth}
  \centering
  \includegraphics[width= 0.65\linewidth]{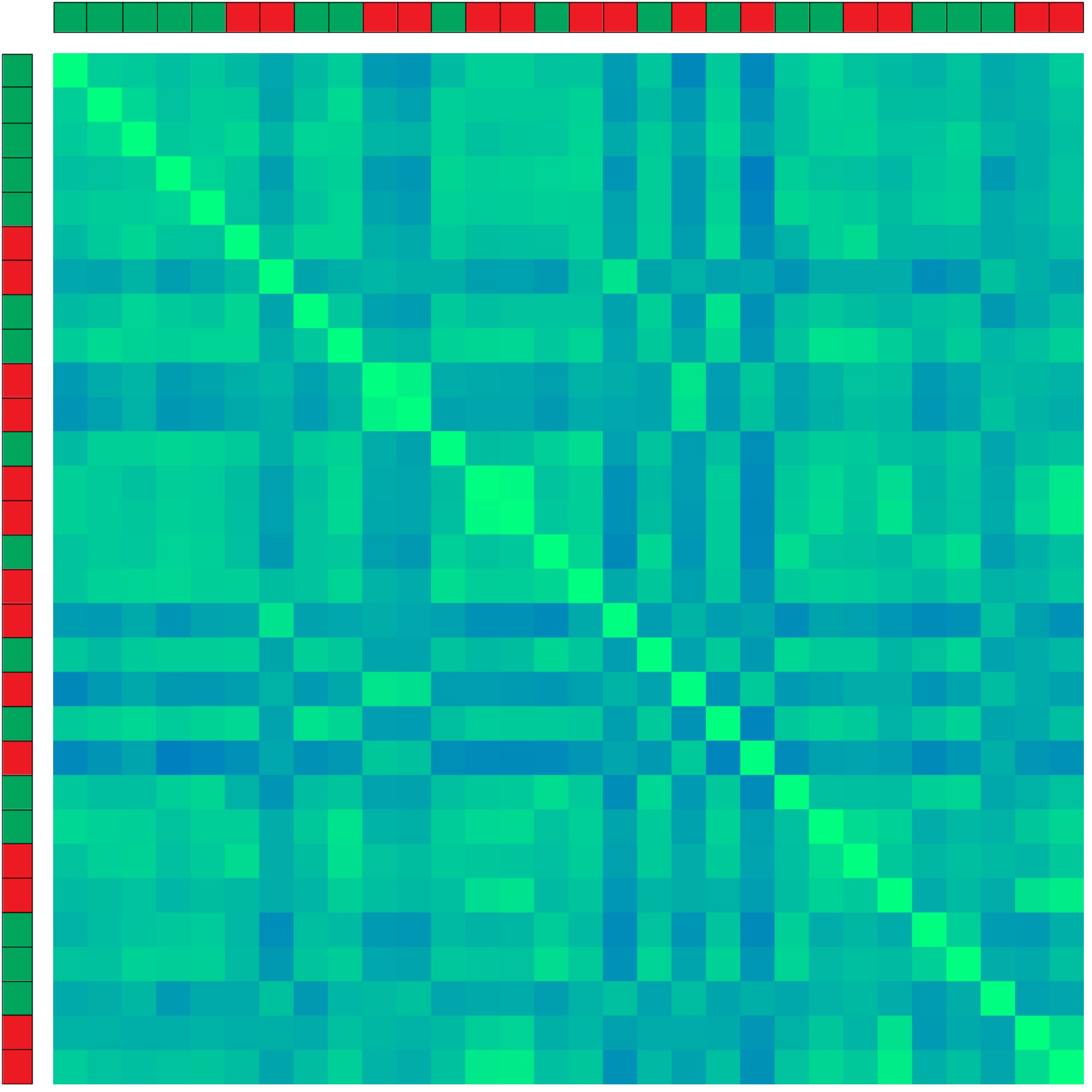}
  \caption{}
  \label{fig:pose}
\end{subfigure}%
\newline
 \vspace{0.1cm}
 \crule[green!50!black!100]{0.2cm}{0.2cm} \footnotesize	{image labeled as \emph{`real'} Melania} \hspace{0cm} \crule[red]{0.2cm}{0.2cm} \footnotesize	{image labeled as \emph{`body double'}}
 \vspace{0.1cm}
\caption{Affinity matrix of cosine similarity between ArcFace embeddings of Melania Trump's images, sorted according to: (a) Occlusion by sunglasses (b) Femininity (c) Head pose.}
\label{fig:face_quality}
\vspace{-0.5cm}
\end{figure}

Figure~\ref{fig:face_quality} portrays the affinity matrices of the cosine similarity scores by sorting the image samples in an ascending order according to occlusion by sunglasses, femininity and head pose quality metrics, in order to highlight their impact on face recognition performance.

In Figures~\ref{fig:sunglasses} and ~\ref{fig:beauty}, one could perceive that the images with low quality scores yield low similarity scores and correspond to the image samples of the \emph{`body double'}, whereas the images with the highest quality coincide with samples of the \emph{`real'} Melania and they deliver relatively higher similarity scores. The linear behaviour observed in Figure~\ref{fig:sunglasses} is a confirmation that sunglasses have high impact on the variation of face recognition scores when comparing the set of \emph{`real'} Melania with the alleged \emph{`body double'}. The two crossed shaped blue areas, in Figure~\ref{fig:beauty}, depicting low similarity scores for higher femininity values correspond to two samples of Melania wearing sunglasses. However, Figure~\ref{fig:pose} shows the pose variation is barely influencing the recognition: comparisons between good and bad quality samples output higher or lower similarity scores alike. This behaviour is an indicator that the used face descriptors are robust enough to the slight head pose variation across Melania's image samples.

Among the selected quality metrics, one could affirm that brightness, face luminance and occlusion by sunglasses are the most relevant quality metrics affecting the face recognition performance, and to a lesser degree with exposure, contrast and femininity. Brightness, face luminance, exposure and contrast being all related to the illumination conditions during the image acquisition depict the same behaviour of the affinity matrix that could be seen in figures~\ref{fig:brightness},~\ref{fig:luminance},~\ref{fig:exposure} and~\ref{fig:contrast}. 
The slight shift between the genuine and imposter score distributions deduced from Melania Trump image samples, observed in subsection~\ref{sec3.3}, is certainly due to the fact that the set of the \emph{`real'} Melania contains overall images of better quality compared to the set of the alleged \emph{`body double'}, and obviously not because of the existence of a Melania Trump \emph{`body double'}.
\section{Conclusion}
In this paper, different state-of-the-art descriptors of face recognition were used to compare the \emph{`real'} Melania image set to the the images of the alleged \emph{`body double'}. According to the proposed study, the answer to the question \emph{`Does Melania Trump have a body double?'} is definitely a \emph{No}. Automatic face recognition invalidates the polemical conspiracy theory claiming the existence of a replacement of Melania Trump during the presidency term of Donald Trump. This disclaimer could only be reliable if we presume that the provided face image samples were not priorly manipulated. Recently, a new type of image manipulation intended to trick the deep leaning based models into delivering an erroneous output. This type of image manipulation, called adversarial attacks, is performed by adding an adversarial perturbation that can be imperceptible to the human eye~\cite{bisogni2021adversarial}. It could be imagined, for instance, that a replacement of Melania Trump exists indeed and that the associated image samples were manipulated in a way that the replacement will be recognized as Melania Trump.

\vspace{0.5cm}

{\small
\bibliographystyle{ieee}
\bibliography{egbib}

\begin{thebibliography}{10}\itemsep=-1pt

\bibitem{mobilesqueezenet}
Mobilesqueezenet: face reidentification retail 0095.
\newblock t.ly/MqxX.

\bibitem{antipov}
G.~Antipov, S.-A. Berrani, and J.-L. Dugelay.
\newblock Minimalistic cnn-based ensemble model for gender prediction from face
  images.
\newblock {\em Pattern recognition letters}, 2016.

\bibitem{becerra2021attribute}
F.~Becerra-Riera, A.~Morales-Gonz{\'a}lez, H.~M{\'e}ndez-V{\'a}zquez, and J.-L.
  Dugelay.
\newblock Attribute-based quality assessment for demographic estimation in face
  videos.
\newblock In {\em 2020 25th International Conference on Pattern Recognition
  (ICPR)}, 2021.

\bibitem{bisogni2021adversarial}
C.~Bisogni, L.~Cascone, J.-L. Dugelay, and C.~Pero.
\newblock Adversarial attacks through architectures and spectra in face
  recognition.
\newblock {\em Pattern Recognition Letters}, 2021.

\bibitem{deng2020retinaface}
J.~Deng, J.~Guo, E.~Ververas, I.~Kotsia, and S.~Zafeiriou.
\newblock Retinaface: Single-shot multi-level face localisation in the wild.
\newblock In {\em Proceedings of the IEEE/CVF Conference on Computer Vision and
  Pattern Recognition}, 2020.

\bibitem{deng2019arcface}
J.~Deng, J.~Guo, N.~Xue, and S.~Zafeiriou.
\newblock Arcface: Additive angular margin loss for deep face recognition.
\newblock In {\em Proceedings of the IEEE/CVF Conference on Computer Vision and
  Pattern Recognition}, pages 4690--4699, 2019.

\bibitem{dolhansky2020deepfake}
B.~Dolhansky, R.~Howes, B.~Pflaum, N.~Baram, and C.~C. Ferrer.
\newblock The deepfake detection challenge (dfdc) dataset.
\newblock {\em arXiv preprint arXiv:2006.07397}, 2020.

\bibitem{howard2017mobilenets}
A.~G. Howard, M.~Zhu, B.~Chen, D.~Kalenichenko, W.~Wang, T.~Weyand,
  M.~Andreetto, and H.~Adam.
\newblock Mobilenets: Efficient convolutional neural networks for mobile vision
  applications.
\newblock {\em arXiv preprint arXiv:1704.04861}, 2017.

\bibitem{liu2017sphereface}
W.~Liu, Y.~Wen, Z.~Yu, M.~Li, B.~Raj, and L.~Song.
\newblock Sphereface: Deep hypersphere embedding for face recognition.
\newblock In {\em Proceedings of the IEEE conference on computer vision and
  pattern recognition}, pages 212--220, 2017.

\bibitem{mahfoudi2019defacto}
G.~Mahfoudi, B.~Tajini, F.~Retraint, F.~Morain-Nicolier, J.~L. Dugelay, and
  P.~Marc.
\newblock Defacto: Image and face manipulation dataset.
\newblock In {\em 27th European Signal Processing Conference (EUSIPCO)}, 2019.

\bibitem{fusionMallat}
K.~Mallat, N.~Damer, F.~Boutros, and J.-L. Dugelay.
\newblock Robust face authentication based on dynamic quality-weighted
  comparison of visible and thermal-to-visible images to visible enrollments.
\newblock In {\em 2019 22th International Conference on Information Fusion
  (FUSION)}, pages 1--8, 2019.

\bibitem{femininity}
K.~Mallat, C.~Galdi, and J.-L. Dugelay.
\newblock Evaluation of the facial makeup impact on femininity appearance based
  on automatic prediction.
\newblock COMET 2017, 2nd Cosmetic Measurement \&amp; Testing Symposium,
  Cergy-Pontoise, France, 2017.

\bibitem{mendez2012evaluacion}
H.~M{\'e}ndez-V{\'a}zquez, L.~Chang, D.~Rizo-Rodr{\'\i}guez, and
  A.~Morales-Gonz{\'a}lez.
\newblock Evaluaci{\'o}n de la calidad de las im{\'a}genes de rostros
  utilizadas para la identificaci{\'o}n de las personas.
\newblock {\em Computaci{\'o}n y Sistemas}, 16(2):147--165, 2012.

\bibitem{PHILLIPS201474}
P.~J. Phillips and A.~J. O'Toole.
\newblock Comparison of human and computer performance across face recognition
  experiments.
\newblock {\em Image and Vision Computing}, 32(1):74--85, 2014.

\bibitem{wu2018light}
X.~Wu, R.~He, Z.~Sun, and T.~Tan.
\newblock A light cnn for deep face representation with noisy labels.
\newblock {\em IEEE Transactions on Information Forensics and Security}, 2018.

\end{thebibliography}
}

\end{document}